\documentclass[runningheads]{llncs}
\usepackage[T1]{fontenc}
\usepackage{graphicx,verbatim}
\usepackage{amsmath}
\usepackage{amssymb}
\usepackage{subfigure}
\usepackage{booktabs} 
\usepackage{multirow} 
\usepackage{tabulary}
\usepackage{url}
\usepackage{microtype}
\usepackage{marvosym}
\begin{document}
\title{Rethinking the Adaptation of Vision Foundation Models for Efficient Cell Segmentation}
\titlerunning{Rethinking the Adaptation of VFMs for Efficient Cell Segmentation}
%

\author{Qing Xu\inst{1,2} \and 
Xiangjian He\inst{2}\textsuperscript{\Letter} \and
Wenting Duan\inst{3} \and
Jiebo Luo\inst{4} \and
Zhen Chen\inst{5}\textsuperscript{\Letter}}

\authorrunning{Q. Xu et al.}

\institute{
School of Computer Science, University of Nottingham, UK \and
School of Computer Science, University of Nottingham Ningbo China, China \and
School of Engineering and Physical Science, University of Lincoln, UK \and
Department of Computer Science, University of Rochester, USA \and
Department of Data Science and Artificial Intelligence,\\The Hong Kong Polytechnic University, Hong Kong SAR 
\\
}
  
\maketitle              
\begin{abstract}
Cell segmentation is critical for computational pathology and biomedical discovery. While recent Vision Foundation Models (VFMs) have demonstrated remarkable universal feature representations, unlocking their full potential for cellular imaging is currently bottlenecked by resource-intensive adaptation paradigms. Existing methods typically rely on fine-tuning heavy visual encoders, leading to extensive computational overhead and a dependency on large-scale annotations. To address this, we propose the EffiCell-Seg framework for highly efficient cell segmentation without re-training the visual encoder. Our core insight is that pretrained VFMs intrinsically encode complementary structural priors: global saliency for localizing potential cells, and local morphological patterns for delineating cellular structures. To harness these priors, we devise a Cell Structure Prompt Encoder (CSP-Encoder) that synthesizes semantic-aware saliency and principal morphological features from frozen VFM representations into explicit structural prior maps. Moreover, we propose a Synergistic Mask Decoder (SM-Decoder) that enforces contextual consistency by jointly predicting geometric distance fields and semantic maps via mutual cross-guidance. Extensive experiments demonstrate that EffiCell-Seg outperforms state-of-the-art methods across diverse cell imaging modalities while requiring only $\sim$5M trainable parameters, over 130$\times$ fewer than fully fine-tuned VFM counterparts. The code is available at \url{https://github.com/xq141839/EffiCell-Seg}.


\end{abstract}
\section{Introduction}
Cell segmentation plays a crucial role in computational pathology and high-throughput biomedical discovery \cite{stringer2021cellpose,li2025nuhtc,xu2025co,xu2025co-journal}. However, this task is inherently challenging due to the diversity of imaging modalities (\textit{e.g.}, H\&E staining, fluorescence microscopy), which exhibit varying contrast ratios and heterogeneous cellular morphologies \cite{lou2025instance,griebel2025segment,lou2025nusegdg}. Recently, Vision Foundation Models (VFMs) such as DINOv3 \cite{simeoni2025dinov3}, pre-trained on massive natural images, have shown remarkable capabilities in capturing universal feature representations, offering a promising avenue to address these issues. While VFMs possess immense potential, unlocking their full capability for cellular imaging is currently bottlenecked by resource-intensive adaptation paradigms, rather than the intrinsic representational limits of the models themselves.


The classical approaches \cite{horst2024cellvit,marks2025cellsam,archit2025segment} fully fine-tuned the entire VFM encoder, incurring substantial computational overhead. To further enhance region-level guidance, \cite{na2024segment,shui2024unleashing} trained auxiliary segmentation or detection models to provide bounding box or point prompts, assisting the VFM in localizing potential cell regions, but at the cost of additional model complexity. Moreover, parameter-efficient fine-tuning techniques \cite{li2026uniultra,zhang2026freqdino} such as Adapter \cite{chenvision} and LoRA \cite{hu2022lora} have emerged as promising alternatives that enable effective adaptation to cell instance segmentation while reducing the number of trainable parameters \cite{nam2024instasam,chen2025sam}.

\begin{figure}[!t]
  \centering
  \includegraphics[width=0.9\linewidth]{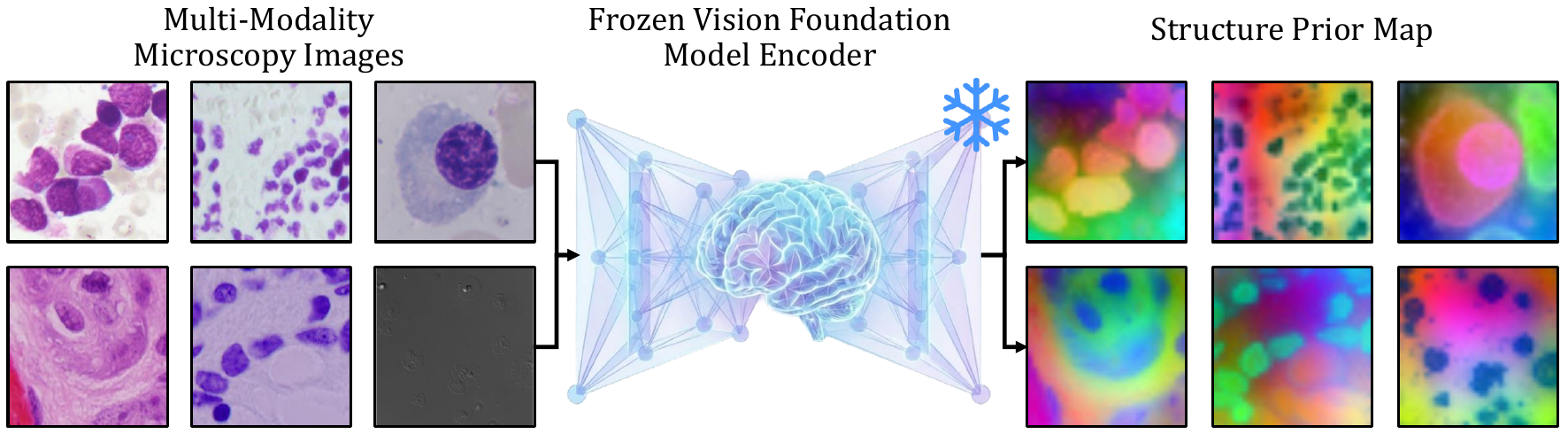}
\caption{Visualization of structural prior maps derived from the pretrained VFM. Given diverse multi-modality microscopy images, the frozen VFM encoder produces structural prior maps that effectively highlight potential cell regions and capture cellular structures.}
  \label{fig:intro}
\end{figure}

Despite these advancements, a fundamental limitation of existing methods \cite{marks2025cellsam,archit2025segment,gao2025dino} is their reliance on re-training the feature representations of the pretrained visual encoder, which typically requires large-scale annotated data. However, acquiring pixel-level cell annotations is notoriously labor-intensive, making it often prohibitive in practice. In fact, VFMs pre-trained on massive natural datasets already possess the capability to perceive universal structural patterns (\textit{e.g.}, textures, boundaries, and spatial continuity), as illustrated in Fig. \ref{fig:intro}. These structural priors are largely shared across natural and cellular images, as cell instances inherently exhibit closed boundaries and locally coherent regions that closely resemble general visual structures. This observation motivates us to rethink the adaptation of VFMs and selectively re-align the existing structural knowledge of pretrained encoders to cellular instance segmentation.

To overcome this bottleneck, we propose the EffiCell-Seg framework for highly efficient cell segmentation that completely circumvents the need to re-train the heavy visual encoder. Our core insight is that the pretrained VFM intrinsically encodes two complementary structural priors: global saliency that highlights potential cell regions, and local morphological patterns that delineate cellular boundaries. Based on this observation, we make the following contributions: (1) We devise a Cell Structure Prompt Encoder (CSP-Encoder) that synthesizes semantic-aware saliency and principal morphological features from frozen VFM representations into explicit structural prior maps, generating high-quality prompts that bridge pretrained knowledge and cellular structures. (2) We further devise a Synergistic Mask Decoder (SM-Decoder) that enforces contextual consistency by jointly predicting geometric distance fields and semantic maps via mutual cross-guidance. To the best of our knowledge, EffiCell-Seg is the first to achieve cell segmentation by exclusively re-aligning the inherent structural priors of a fully frozen VFM. (3) Extensive experiments show that EffiCell-Seg achieves state-of-the-art performance across diverse imaging modalities. Remarkably, it achieves such superior accuracy while requiring only $\sim$5M trainable parameters, over $130\times$ fewer than fully fine-tuned VFM counterparts.


\section{Methodology}

\begin{figure}[!t]
  \centering
  \includegraphics[width=0.94\linewidth]{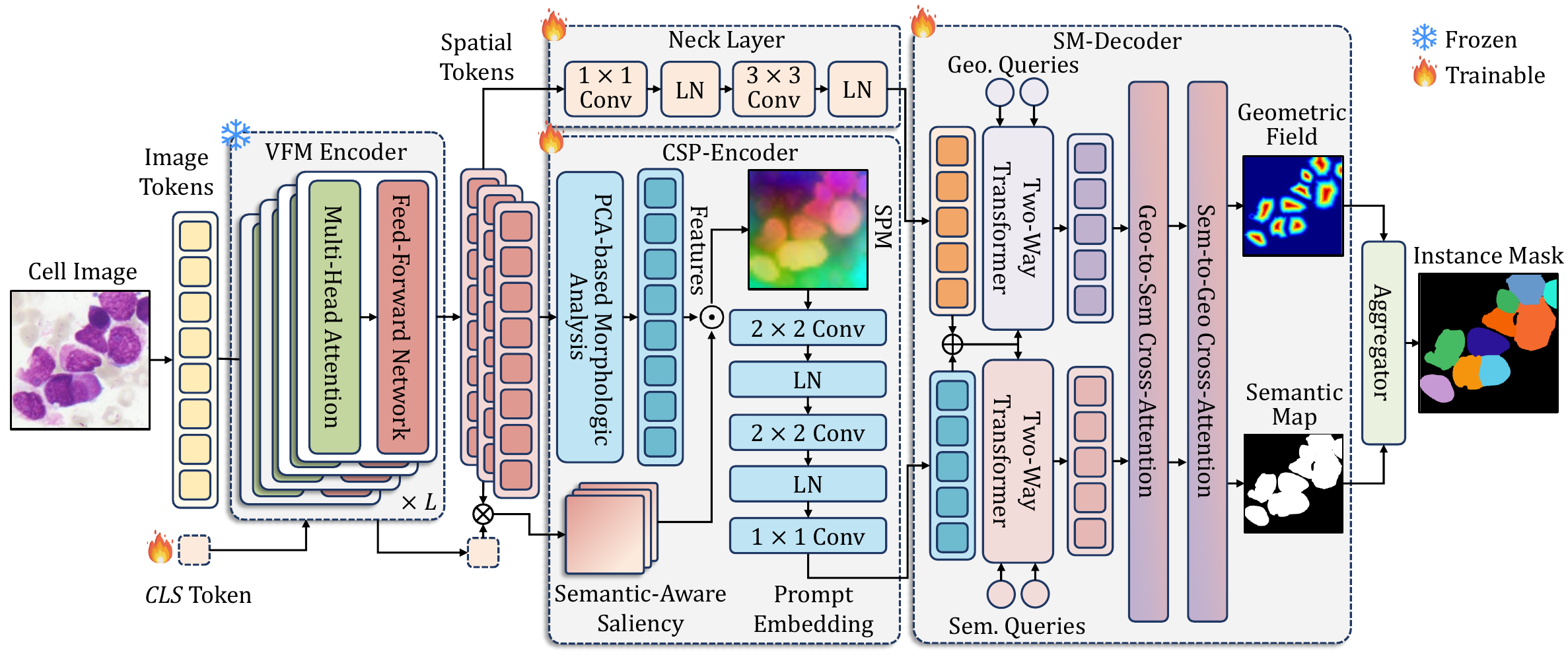}
\caption{Overview of the proposed EffiCell-Seg framework. The CSP-Encoder fuses semantic-aware saliency and PCA-based morphology from pretrained VFM features into structural prior maps, generating high-quality prompts for the SM-Decoder to jointly infer cell geometric structure and semantic map through cross-guided decoding.}
  \label{fig:method}
\end{figure}

As illustrated in Fig. \ref{fig:method}, we propose the EffiCell-Seg framework that unlocks the structural knowledge of pretrained VFMs for efficient cell segmentation without re-training the heavy visual encoder. Specifically, we devise a CSP-Encoder that synthesizes semantic-aware saliency and morphological features from frozen VFM representations into structural prior maps, bridging pretrained knowledge and cellular structures. We further design an SM-Decoder that jointly predicts geometric distance fields and semantic maps via mutual cross-guidance, ensuring contextual consistency for accurate segmentation.

\subsection{Preliminary and Insights}
Current VFMs such as DINOv3 \cite{simeoni2025dinov3} typically adopt Vision Transformer (ViT) as the visual encoder, pre-trained on large-scale natural image datasets. Given an input image $x$, we partition it into $N$ non-overlapping patches and feed them into the pretrained VFM encoder. The encoder consists of $L$ stacked transformer blocks, where each block applies multi-head self-attention (MHSA) and a feed-forward network (FFN) with layer normalization (LN) and residual connections:
\begin{equation}
h_l' = \text{MHSA}(\text{LN}(h_{l-1})) + h_{l-1}, \quad h_l = \text{FFN}(\text{LN}(h_l')) + h_l',
\end{equation}
where $h_0 = [t_{\text{cls}}; t_1, t_2, \ldots, t_N]$ denotes the initial token sequence. After $L$ layers, we obtain the output representations $h_L = \{t_{\text{cls}}, t_1, t_2, \ldots, t_N\}$, where $t_{\text{cls}} \in \mathbb{R}^{D}$ is the CLS token and $\{t_i\}_{i=1}^{N} \in \mathbb{R}^{N \times D}$ are the spatial tokens.

Through self-attention across $L$ layers, $t_{\text{cls}}$ progressively aggregates information from all spatial tokens, forming a holistic summary of the entire image. Its affinity with individual spatial tokens thus naturally reflects the structural saliency of each local region, indicating \textit{where} potential cell regions reside. Meanwhile, the spatial tokens $\{t_i\}_{i=1}^{N}$ preserve rich patch-level information encoding morphological patterns, characterizing \textit{what} cellular structures look like. These two structural priors are inherently complementary: global saliency offers region-level localization while local patterns provide fine-grained structural details. As illustrated in Fig.~\ref{fig:intro}, this insight motivates our EffiCell-Seg, which extracts and fuses these priors into structural prior maps for guiding cell instance segmentation.

\subsection{Cell Structure Prompt Encoder}
Following our insight that the CLS token and spatial tokens of pretrained VFMs encode complementary structural priors, we devise the CSP-Encoder to extract and fuse these priors into structure-aware prompts without any encoder fine-tuning. Specifically, to capture \textit{where} potential cells reside, we compute the cosine similarity between $t_{\text{cls}}$ and each spatial token $t_i$ as a saliency map:
\begin{equation}
a_i = \frac{t_i \cdot t_{\text{cls}}^{\top}}{\|t_i\| \|t_{\text{cls}}\|},
\end{equation}
which effectively highlights foreground cell regions while suppressing background noise. To characterize \textit{what} cellular structures look like, we apply PCA to $\{t_i\}_{i=1}^{N}$ and retain the top-$K$ components as follows:
\begin{equation}
m_i = \text{PCA}_K(\{t_i\}_{i=1}^{N}) \in \mathbb{R}^{K}.
\end{equation}
The principal components capture dominant \textit{morphological} and \textit{textural} variations, encoding fine-grained structural details such as cell boundaries and internal textures. We then fuse these two priors through element-wise multiplication to obtain the structural prior map (SPM) as $z_i = a_i \cdot m_i$, where the saliency score acts as a spatial gate that selectively activates morphological features in structurally relevant regions. Finally, we employ a lightweight convolutional block to transform the structural prior map into the prompt embedding $z$, enforcing local spatial consistency for downstream dense prediction. In this way, the CSP-Encoder bridges frozen VFM representations and cellular morphology, providing reliable spatial guidance for subsequent geometric and semantic decoding.

\subsection{Synergistic Mask Decoder}
The CSP-Encoder provides structure-aware prompt embedding $z$ from frozen VFM features. To achieve accurate cell instance segmentation, we further need to coherently decode this structural prior into dense predictions. To this end, we devise the SM-Decoder that jointly predicts geometric distance fields and semantic maps through cross-guided decoding. The core motivation is that geometric cues (\textit{i.e.}, distance fields) provide precise boundary localization and instance separation but are sensitive to noise in low-contrast regions, while semantic confidence suppresses spurious responses but lacks the spatial precision to separate touching cells. Treating them independently often leads to inconsistent or fragmented predictions. The SM-Decoder therefore enforces mutual structural consistency between them through bidirectional cross-guidance.
Specifically, the SM-Decoder maintains two sets of learnable query embeddings: geometric queries $q_g$ for capturing spatial organization and instance centrality, and semantic queries $q_s$ for encoding region-level category-aware features. We first adopt a projection layer ${\rm Prj}(\cdot)$ to align the dimensions of frozen VFM features with the prompt embedding, and then employ a two-way transformer \cite{kirillov2023_sam}, denoted as ${\rm TWA}(\cdot)$, to initialize geometric and semantic-aware representations as follows:
\begin{equation}
h_g = \text{TWA}(q_g, q_g, {\rm Prj}(h) \oplus z), \quad h_s = \text{TWA}(q_s, q_s, {\rm Prj}(h) \oplus z),
\end{equation}
where $\oplus$ denotes element-wise addition. This grounds both query sets in the same structure-aligned feature space. We further leverage cross-guidance attention to facilitate bidirectional interaction between geometric and semantic embeddings:
\begin{equation}
h_g \gets \text{CrossAttn}(q_g, q_g, h_g \oplus h_s), \quad h_s \gets \text{CrossAttn}(q_s, q_s, h_g \oplus h_s),
\end{equation}
which ensures that geometric precision and semantic confidence mutually reinforce each other, preventing degenerate solutions where either dominates in isolation. Finally, the SM-Decoder generates the distance fields and semantic maps through lightweight transposed convolutions. Through this design, the SM-Decoder coherently decodes structure-aware prompts into geometrically precise and semantically consistent predictions, enabling accurate cell instance segmentation.

\subsection{Optimization Pipeline}

In the design of our EffiCell-Seg framework, we adopt a pretrained DINOv3 \cite{simeoni2025dinov3} as the VFM encoder and freeze its weights throughout training to preserve the universal structural knowledge learned from large-scale natural datasets. This design not only reduces computational overhead but also prevents overfitting to domain-specific appearance variations. Adaptation to cellular scenarios is achieved exclusively through optimizing the lightweight CSP-Encoder and SM-Decoder. During the forward pass, frozen VFM features are first processed by the CSP-Encoder to generate the structure-aware prompt embedding, which then guides the SM-Decoder to jointly infer the horizontal and vertical distance fields and the semantic map through cross-guided decoding.

The training is supervised by a joint loss combining cross-entropy loss $\mathcal{L}_{\mathrm{CE}}$ with dice loss $\mathcal{L}_{\mathrm{Dice}}$ for semantic prediction, and MSE loss $\mathcal{L}_{\mathrm{MSE}}$ for accurate distance estimation. The overall objective is formulated as follows:
\begin{equation}
    \mathcal{L}_{\rm Cell} = \mathcal{L}_{\mathrm{CE}} + \mathcal{L}_{\mathrm{Dice}} + \lambda_1\mathcal{L}_{\mathrm{MSE}},
\end{equation}
where $\lambda_1$ is the balancing coefficient that compensates for the magnitude discrepancy between regression and classification losses \cite{graham2019hover}. During inference, we apply a watershed algorithm \cite{roerdink2000watershed} on the predicted distance field, guided by the semantic map, to separate individual cell instances. By optimizing $\mathcal{L}_{\rm Cell}$, our EffiCell-Seg effectively activates and re-aligns the universal structural knowledge of the frozen VFM for accurate cell segmentation with superior computational efficiency.

\section{Experiments}

\subsection{Experimental Setup}
To validate the effectiveness of the proposed EffiCell-Seg, we conduct experiments on two benchmarks: the CellSeg dataset \cite{ma2024multimodality}, containing 1,101 multimodal microscopy images, and the DSB dataset \cite{caicedo2019nucleus}, comprising 670 images. These two datasets collectively cover diverse cellular imaging modalities and cell types, including plasma cells for Multiple Myeloma and white blood cells for Leukaemia. Both datasets adopt a common split of training, validation, and test sets as 7:1:2. We perform all experiments on a NVIDIA H20 GPU using PyTorch. We adopt DINOv3 \cite{simeoni2025dinov3} as the VFM encoder, whose weights remain frozen throughout training. All input images are resized to $512 \times 512$. We apply the AdamW optimizer with a learning rate of $1\times10^{-4}$ and use CosineAnnealingLR for the scheduling strategy. The batch size and training epochs are set to 16 and 100, respectively. The balancing coefficients $\lambda_1$ are set to 10. For fair comparisons, all baselines are fine-tuned on the same training set and evaluated under the same input resolution. All SAM-based baselines \cite{marks2025cellsam,shui2024unleashing,horst2024cellvit,na2025segment} adopt their automatic prompting mode. For quantitative evaluation, we adopt Dice and mIoU to measure semantic segmentation overlap, and F1 and Panoptic Quality (PQ) to assess instance-level detection and segmentation quality.

\begin{table}[!t]
\centering
\tabcolsep=0.4cm
\caption{Comparison with state-of-the-art methods on cell semantic segmentation.}
\resizebox{\textwidth}{!}{\begin{tabular}{l|c|cc|cc}
\hline
\multirow{2}{*}{Methods} & Trainable & \multicolumn{2}{c|}{ CellSeg } & \multicolumn{2}{c}{ DSB } \\
\cline{3-6}
& \#Params & Dice & mIoU & Dice & mIoU \\
\hline
CPPNet \cite{chen2023cpp} & 32.67M & 84.13 & 74.43 & 90.57 & 84.19  \\
Swin-UMamba \cite{liu2024swin} & 27.43M & 85.24 & 75.81 & 91.23 & 85.02    \\
\hline
CellSAM \cite{marks2025cellsam} & 682.92M & 85.87 & 76.52 & 91.58 & 85.46    \\
PromptNucSeg \cite{shui2024unleashing}  & 661.74M & 86.14 & 76.93 & 91.85 & 85.73    \\
CellViT \cite{horst2024cellvit} & 68.56M & 86.48 & 77.12 & 92.04 & 85.96    \\
SAC~\cite{na2025segment} & 51.83M & 86.92 & 77.51 & 92.31 & 86.34    \\
\hline
EffiCell-Seg (Ours)  & \textbf{4.99M} & \textbf{87.70} & \textbf{78.48} & \textbf{93.14} & \textbf{87.37} \\
\hline
\end{tabular}}
\label{tab:semantic}
\end{table}

\begin{table}[!t]
\centering
\tabcolsep=0.40cm
\caption{Comparison with state-of-the-art methods on cell instance segmentation.}
\resizebox{\textwidth}{!}{\begin{tabular}{l|c|cc|cc}
\hline
\multirow{2}{*}{Methods} & Trainable & \multicolumn{2}{c|}{ CellSeg } & \multicolumn{2}{c}{ DSB } \\
\cline{3-6}
& \#Params & PQ  & F1  & PQ  & F1  \\
\hline
CPPNet \cite{chen2023cpp} & 32.67M & 55.83 & 54.96 & 70.12 & 68.75    \\
Swin-UMamba \cite{liu2024swin} & 27.43M & 56.74 & 55.81 & 71.28 & 69.83    \\
\hline
CellSAM \cite{marks2025cellsam} & 682.92M & 57.92 & 57.13 & 72.17 & 70.92  \\
PromptNucSeg \cite{shui2024unleashing}  & 661.74M & 58.31 & 57.52 & 72.85 & 71.43    \\
CellViT \cite{horst2024cellvit} & 68.56M & 58.65 & 57.94 & 73.24 & 71.86    \\
SAC~\cite{na2025segment} & 51.83M & 59.78 & 58.96 & 74.05 & 72.54   \\
\hline
EffiCell-Seg (Ours)  & \textbf{4.99M} & \textbf{61.16} & \textbf{60.34} & \textbf{75.32} & \textbf{73.89}    \\
\hline
\end{tabular}}
\label{tab:inst}
\end{table}

\begin{figure}[!t]
  \centering
  \includegraphics[width=0.9\linewidth]{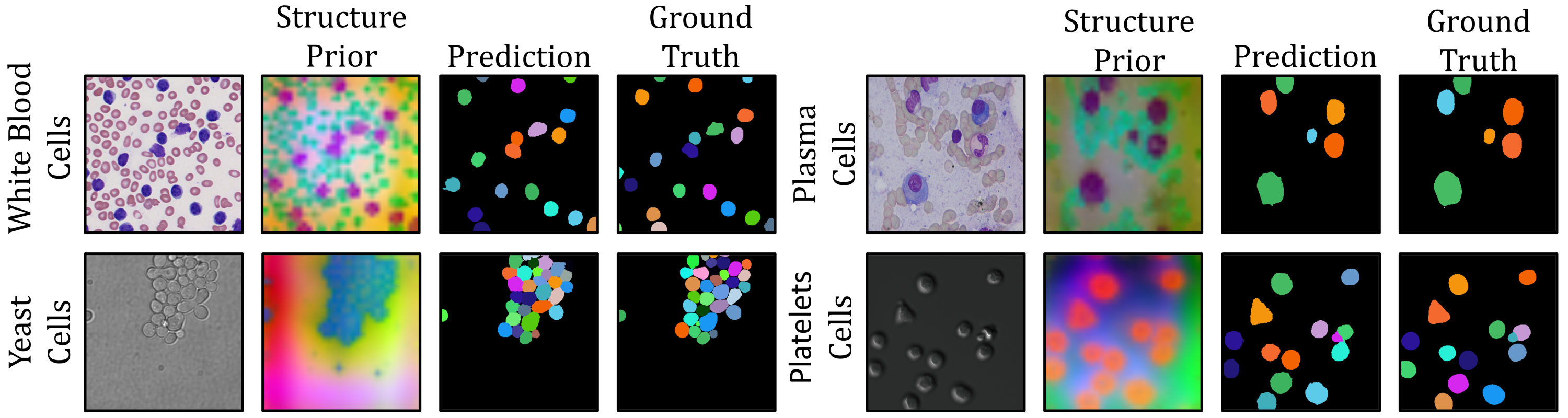}
\caption{Qualitative results of EffiCell-Seg on diverse cell types. By effectively leveraging structural priors from the pretrained VFM, our model achieves precise boundary delineation and robust instance separation across diverse imaging modalities.}
  \label{fig:visual}
\end{figure}

\subsection{Comparison with State-of-the-art Methods}
\noindent \textbf{Cell Semantic Segmentation.} We first evaluate all methods on cell semantic segmentation. As shown in Table~\ref{tab:semantic}, classical approaches \cite{chen2023cpp,liu2024swin} show limited performance due to the lack of large-scale pretrained knowledge. Among VFM-based methods, CellSAM \cite{marks2025cellsam} and PromptNucSeg \cite{shui2024unleashing} achieve moderate results but require over 660M trainable parameters owing to fine-tuning the heavy visual encoder or training auxiliary models. CellViT \cite{horst2024cellvit} and SAC \cite{na2025segment} reduce the parameter cost while improving performance, yet still depend on encoder fine-tuning. Our EffiCell-Seg achieves the best results on both datasets, outperforming the second-best SAC by 0.78\% Dice and 0.97\% mIoU on CellSeg, and 0.83\% Dice and 1.03\% mIoU on DSB, while requiring only \textbf{4.99M} trainable parameters, approximately \textbf{10$\times$ fewer} than SAC and over \textbf{130$\times$ fewer} than CellSAM. These results demonstrate that re-aligning structural knowledge of pretrained VFMs is more effective than re-training the heavy visual encoder for cell segmentation.

\noindent \textbf{Cell Instance Segmentation.} We further evaluate all methods on cell instance segmentation. As shown in Table~\ref{tab:inst}, the classical methods \cite{chen2023cpp,liu2024swin} struggle in instance-level prediction, while heavily parameterized VFM-based methods such as CellSAM \cite{marks2025cellsam} and PromptNucSeg \cite{shui2024unleashing} do not necessarily translate to better instance discrimination. Our EffiCell-Seg achieves the best performance across both datasets, surpassing SAC by 1.27\% PQ and 1.35\% F1 on DSB, demonstrating that the cross-guided decoding in our SM-Decoder effectively leverages geometric and semantic complementarity for coherent instance predictions. The qualitative results in Fig.~\ref{fig:visual} further illustrate that EffiCell-Seg produces highly precise boundary delineation and clear instance separation in challenging regions with overlapping and morphologically diverse cells. These results explicitly validate our core insight: rather than expensively re-training the entire vision foundation model, strategically re-aligning its inherent structural priors via our EffiCell-Seg is a substantially more efficient and robust paradigm for cellular analysis.

\subsection{Ablation Study}
To validate the effectiveness of the proposed CSP-Encoder $\mathcal{P}$ (\textit{i.e.}, including semantic-aware saliency $\alpha$ and PCA Map $m$), and SM-Decoder $\mathcal{D}$, we conduct ablation studies on both CellSeg and DSB datasets for cell segmentation, as shown in Table~\ref{tab:ablation}. Starting from the baseline that directly decodes frozen VFM features without any proposed component, introducing the Similarity Map alone yields 1.89\% Dice and 1.73\% PQ improvements on CellSeg. Adding the PCA Map alone brings comparable gains of 1.41\% Dice and 1.26\% PQ. Incorporating a complete CSP-Encoder achieves further improvements across all metrics, demonstrating that this prior provides reliable region localization while local morphological patterns offer fine-grained structural details. Finally, incorporating the SM-Decoder with cross-guided decoding brings the most significant boost of 1.42\% Dice and 1.63\% PQ on CellSeg and 1.20\% Dice and 1.46\% PQ on DSB, highlighting the importance of enforcing geometric-semantic consistency for both semantic and instance-level predictions. These comprehensive results validate that the tailored CSP-Encoder and SM-Decoder collectively contribute to the superior performance of our framework across diverse cell segmentation scenes.

\begin{table}[!t]
\centering
\tabcolsep=0.18cm
\caption{Ablation study of our EffiCell-Seg framework on cell segmentation.}
\resizebox{\textwidth}{!}{\begin{tabular}{cc|c|cccc|cccc}
\hline
\multicolumn{2}{c|}{$\mathcal{P}$} & \multirow{2}{*}{$\mathcal{D}$} & \multicolumn{4}{c|}{CellSeg} & \multicolumn{4}{c}{DSB} \\
\cline{1-2} \cline{4-11}
$\alpha$ & $m$ & & Dice  & mIoU  & PQ  & F1  & Dice  & mIoU  & PQ  & F1  \\
\hline
 &  &  & 83.52 & 73.86 & 56.42 & 55.63 & 89.74 & 83.15 & 71.05 & 69.48 \\
\checkmark &  &  & 85.41 & 75.92 & 58.15 & 57.26 & 91.26 & 84.83 & 72.74 & 71.23 \\
 & \checkmark &  & 84.93 & 75.48 & 57.68 & 56.84 & 90.87 & 84.41 & 72.31 & 70.85 \\
\checkmark & \checkmark &  & 86.28 & 76.85 & 59.53 & 58.71 & 91.94 & 85.72 & 73.86 & 72.34 \\
\checkmark & \checkmark & \checkmark & \textbf{87.70} & \textbf{78.48} & \textbf{61.16} & \textbf{60.34} & \textbf{93.14} & \textbf{87.37} & \textbf{75.32} & \textbf{73.89} \\
\hline
\end{tabular}}
\label{tab:ablation}
\end{table}

\section{Conclusion}
We present EffiCell-Seg, a highly \textit{efficient} and \textit{effective} cell segmentation framework that circumvents heavy visual encoder fine-tuning. Our core insight is that pretrained VFMs intrinsically encode complementary structural priors for cellular analysis. To rethink the VFM adaptation, we devise the CSP-Encoder to synthesize semantic-aware saliency and principal morphological features into structural prior maps. Moreover, we propose the SM-Decoder to jointly infer geometric distance fields and semantic maps via cross-guided decoding, ensuring robust contextual consistency. Extensive evaluations demonstrate that the proposed EffiCell-Seg framework outperforms state-of-the-art methods across diverse imaging modalities, delivering superior segmentation accuracy while requiring much fewer trainable parameters than existing VFM adaptation paradigms.



\bibliographystyle{splncs04}
\bibliography{refs}

\end{document}